\documentclass[11pt]{article}

\usepackage[preprint]{acl}

\usepackage[T1]{fontenc}
\usepackage[utf8]{inputenc} %
\usepackage{times}
\usepackage{latexsym}
\usepackage{hyperref}
\usepackage{float}

\usepackage[utf8]{inputenc}

\usepackage{microtype}

\usepackage{inconsolata}

\usepackage{amsmath}
\usepackage{microtype}
\usepackage{hyperref}
\usepackage{url}
\usepackage{booktabs}
\usepackage{graphicx}
\usepackage{multirow}
\usepackage{wrapfig} %

\usepackage{lineno}

\definecolor{darkblue}{rgb}{0, 0, 0.5}
\hypersetup{colorlinks=true, citecolor=darkblue, linkcolor=darkblue, urlcolor=darkblue}

\usepackage{lipsum}
\usepackage{amssymb}

\usepackage {xcolor} 
\usepackage {listings}

\lstdefinestyle{json}{
    basicstyle=\ttfamily\footnotesize,
    commentstyle=\color{gray},
    numberstyle=\tiny\color{gray},
    stepnumber=1,
    numbersep=5pt,
    backgroundcolor=\color{white},
    showspaces=false,
    showstringspaces=false,
    showtabs=false,
    frame=single,
    rulecolor=\color{black},
    tabsize=2,
    breaklines=true,
    breakatwhitespace=true,
    captionpos=b,
    keywordstyle=\color{blue}\bfseries,
    stringstyle=\color{red},
    emphstyle=\color{magenta}\itshape,
    morecomment=[l]{//},
    morekeywords={id, image, gt, data_source, conversations, filter, judge_response},
    emph={unique_identifier, image_path, ground_truth, dataset_source, question, question8, cot, valid, invalid, filter_reason}
}

\title{\smash{\raisebox{-0.35em}{\includegraphics[height=1.5em]{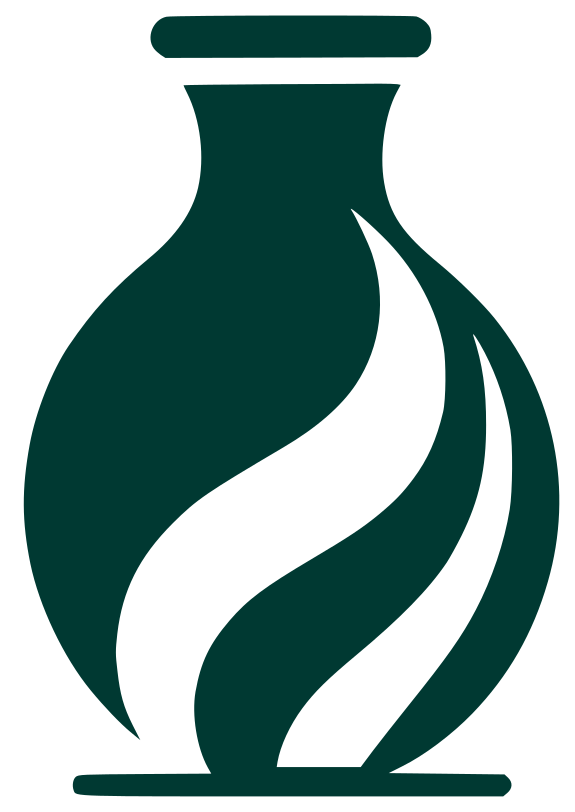}}~VaseVQA: Multimodal Agent and Benchmark for Ancient Greek Pottery}}

\author{Jinchao Ge$^{1*}$~%
Tengfei Cheng$^{2*}$~
Biao Wu$^{3*}$~
Zeyu Zhang$^{4*\dag}$~ %
Shiya Huang$^{1}$~%
Judith Bishop$^{5}$\\ %
\textbf{Gillian Shepherd}$^{5}$ ~ %
\textbf{Meng Fang}$^{2}$ ~ %
\textbf{Ling Chen}$^{3}$ ~ %
\textbf{Yang Zhao}$^{5\ddag}$ \vspace{0.3em}\\ %
$^{1}$University of Adelaide~ $^{2}$University of Liverpool ~ $^{3}$University of Technology Sydney \\~ $^{4}$The Australian National University $^{5}$La Trobe University\\
\small$^{*}$Equal contribution $^{\dag}$Project lead $^{\ddag}$Corresponding author: y.zhao2@latrobe.edu.au}

\begin{document}
\maketitle

\begin{abstract}

Understanding cultural heritage artifacts such as ancient Greek pottery requires expert-level reasoning that remains challenging for current MLLMs due to limited domain-specific data. We introduce VaseVQA, a benchmark of 31{,}773 images and 67{,}614 question--answer pairs across seven expert-defined categories, enabling systematic evaluation of expert-level cultural heritage understanding. Using this dataset, we explore effective training strategies for domain-specific reasoning. While supervised fine-tuning improves adaptation to domain knowledge, it struggles with deeper reasoning tasks. We propose VaseVL, which augments SFT with reinforcement learning using verifiable rewards. Experiments show that VaseVL consistently outperforms supervised baselines, especially on reasoning-intensive questions, highlighting the value of targeted reinforcement learning for cultural heritage visual question answering. Our code and dataset will be released at \url{https://github.com/AIGeeksGroup/VaseVQA}.
\end{abstract}

\begin{figure*}[t]
    \centering
    \includegraphics[width=0.9\linewidth]{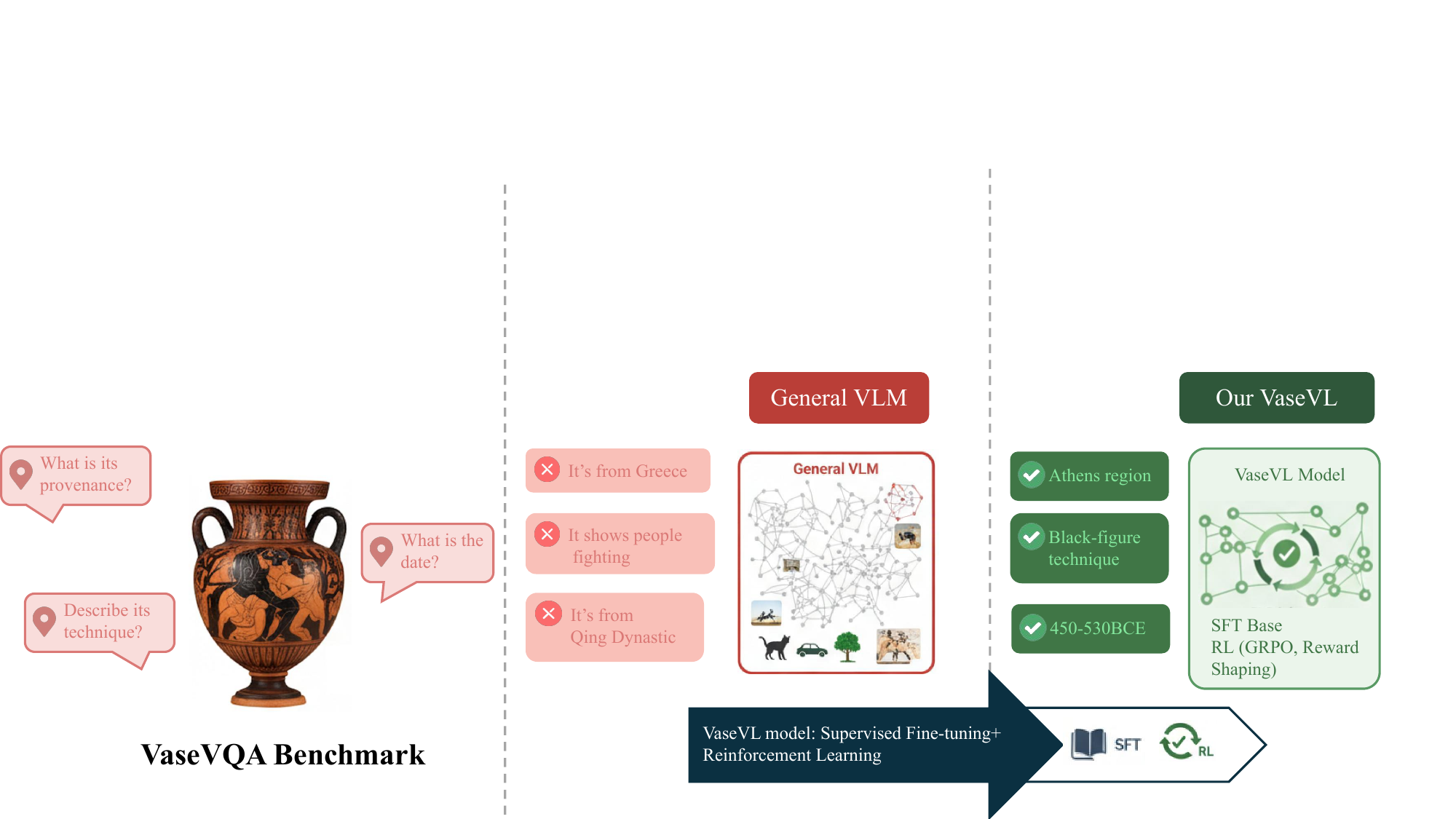}
    \vspace{-3mm}
    \caption{Bridging the Semantic Gap: Limitations of Existing Vision-Language Models on Ancient Greek Vase Understanding and the Proposed VaseVL Framework}
    \label{fig:teaser}
    \vspace{-1em}
\end{figure*}

\section{Introduction}
Cultural heritage artifacts are physical objects that connect us to the past and provide valuable information about art, technology, and society over time. Computational analysis of these artifacts has been a longstanding objective in digital humanities, aiming to augment expert knowledge, facilitate large-scale cataloging, and democratize access to specialized expertise~\citep{Castellano_2022}. While recent advances in vision-language models have demonstrated remarkable capabilities in general-domain tasks~\citep{li2023blip2bootstrappinglanguageimagepretraining, liu2023visualinstructiontuning}, their applicability to specialized domains that demand deep expert knowledge remains constrained. Ancient Greek pottery is a good example of why analyzing cultural heritage artifacts is challenging. A single vase can provide different types of information: the clay reveals where it was made, the shape suggests its use and time period, the decoration reflects workshop traditions, and stylistic details help identify specific artists or schools~\citep{smith202422}. Domain experts synthesize these visual cues with extensive contextual knowledge, ranging from regional manufacturing practices to historical timelines and scholarly conventions, to answer complex questions about where the objects come from, when they were made, and who made them.  Automating this analysis requires models that can both recognize fine visual details and reason using specialized domain knowledge—capabilities that go beyond what current general-purpose MLLMs can provide.

However, most existing vision-language models lack the domain-specific knowledge required to answer these fundamental questions reliably. Such knowledge is inherently scarce in the real world, as it is typically confined to expert literature, museum archives, and specialized scholarship. As a result, current VLMs are underexposed to this type of information during pre-training, which limits their ability to reason about cultural heritage artifacts beyond surface-level visual cues. To address this gap, we curate and construct a dedicated dataset that systematically captures expert knowledge for ancient Greek pottery, providing a foundation for training and evaluating models on these core archaeological questions. To facilitate systematic evaluation, we introduce VaseVQA, a comprehensive benchmark comprising 31{,}773 images (featuring an 11{,}693-image single-view subset) and 67{,}614 visual question–answer pairs. The benchmark encompasses seven distinct question types and incorporates type-specific evaluation metrics, enabling rigorous assessment of both lexical precision and semantic alignment.

With the dataset in place, a natural question is how existing vision–language models can be adapted to reason over such expert knowledge. Since the annotations in VaseVQA lie largely outside the pre-training distribution of current MLLMs, standard supervised fine-tuning can partially bridge this gap by improving factual recall. However, our empirical analysis shows that exposure to expert data alone is insufficient for robust reasoning over attributes that are not directly observable, such as provenance, dating, or attribution. This observation motivates the use of additional training signals beyond conventional supervision to better align models with expert-level reasoning requirements.~\citep{huang2025visionr1incentivizingreasoningcapability,wang2025thinknotselectivereasoning}. We define a seven-category question taxonomy—\emph{Fabric}, \emph{Technique}, \emph{Shape}, \emph{Provenance}, \emph{Attribution}, \emph{Date}, and \emph{Decoration}—and use it to identify type-specific weaknesses in post-SFT models. Based on this taxonomy, we design a rule-based reward to guide reinforcement learning, aiming to improve performance on more challenging questions by strengthening the model’s reasoning ability.

\begin{table*}[!t]
    \centering
    \resizebox{\textwidth}{!}{
        \begin{tabular}{l|cccccccc} \toprule
            Datasets & Images & Questions & Question Type & Image Type & Task Focus & Venue&OE/MC\\
            \midrule
            DAQUAR\cite{malinowski2014multi} & 1,449 & 12,468 &4&Natural&VQA&NIPS 2014&OE   \\
            COCO-QA\cite{ren2015exploring} & 123,287& 117,684 &4&Natural&VQA&-&OE\\
            VAQ V1.0\cite{agrawal2017c} & 204k & 614K &-&Natural&VQA&ICCV 2015&OE\\
            VQA V2.0\cite{goyal2017making} & 204k & 1.1M &-&Natural&VQA&CVPR 2017&Both\\
            CVR\cite{zellers2019recognition} & 110k & 290K&&Natural&VR&CVPR 2019&MC\\ 
            GQA\cite{hudson2019gqa} & 113,018 &22,669,678&-&Natural&VR&CVPR 2019&OE\\ 
            RAVEN\cite{zhang2019raven} &1,120,000 &70,000 &4& Natural&VR&CVPR 2019&MC\\
            NLVR\cite{suhr2017corpus} &387,426&31,418&-&Synthetic&VR&ACL 2019&OE\\
            OK-VQA\cite{marino2019ok} &14,031&14,055&VQA&Natural&VQA&CVPR 2019&OE\\
            VizWiz\cite{gurari2018vizwiz} &-&31,173&-&Natural&-&CVPR 2018&OE\\
            KVQA\cite{shah2019kvqa} & 24K&&&Natural&-&AAAI 2019&OE\\ 
            CLEVR\cite{johnson2017clevr} &100,000&999,968&90&Natural&VR&CVPR 2017&OE\\ 
            FM-IQA\cite{gao2015you} &158,392&316,193&–&Natural& - &CVPR 2017&OE\\
            NLVR2\cite{suhr2019nlvr2} &107,292&29,680&-&Synthetic&VR&ACL 2019&OE\\
            TextVQA\cite{singh2019towards} &28,408&45,336&-&Natural&VQA&CVPR 2019&OE\\
            FVQA\cite{wang2017fvqa} &2190&5826&12&Natural&VR&CVPR 2019&OE\\
            VISUAL GENOME\cite{krishna2017visual} &108,000&145,322&7&Natural&VR&-&OE\\ 
            VQA-CP\cite{agrawal2018don} &&&&Natural&VQA&CVPR 2018\\
            Visual Madlibs\cite{yu2015visual} &10,738&360,001&12&Natural&VR&-&-\\ 
            SHAPES\cite{andreas2015deep} &15,616&244&–&Synthetic&VR &-&Binary\\ 
            KB-VQA\cite{wang2015explicit} &700&2402&23&Natural& VQA&IJCAI 17&OE\\
            ICQA\cite{hosseinabad2021multiple} &42,021&260,840&-&Synthetic&VQA &-&OE\\
            DVQA\cite{kafle2018dvqa} &3,000,000&3,487,194&3&-&VQA &-&OE\\
            PathVQA\cite{he2020pathvqa} &4,998&32,795&7&-&VQA &-&MC\\
            Visual7w\cite{zhu2016cvpr} &47,300&327,939&7&Natural&VQA &CVPR 16&MC\\
            KRVQA\cite{cao2021knowledge} &32,910&157,201&6&Natural&VR &-&MC\\
            \midrule
            VaseVQA & 11,693 & 67,614 & 7 & Natural & VQA & - & OE \\
            \bottomrule
        \end{tabular}
    }
    \caption{A Main characteristics of major VQA and Visual Reasoning datasets.} \label{tab:major_VQA_datasets} 
    \vspace{-3mm}
\end{table*}

Experimental results show that general-purpose MLLMs still face a clear domain gap in cultural heritage understanding. Although these models achieve strong baseline performance on general tasks, they struggle with expert-level questions in zero-shot settings, largely due to the scarcity of domain-specific knowledge and the lack of specialized training data for artifacts such as Ancient Greek pottery. SFT substantially improves factual recall, achieving near-ceiling performance on visually explicit categories (e.g., \emph{Fabric} and \emph{Technique}), but remains unstable on reasoning-intensive questions, indicating its limited ability to model deeper contextual relationships. In contrast, VaseVL addresses these limitations by introducing verifiable rewards that explicitly supervise reasoning outcomes, consistently outperforming the SFT baseline on challenging question types while also improving compositional robustness.

Our main contributions are as follows:
\begin{itemize}
\item We introduce VaseVQA, a benchmark comprising 31{,}773 images and 67{,}614 question–answer pairs. This work fills a critical gap in the field by providing a large-scale, expert-annotated dataset for Ancient Greek vases and porcelain, covering seven expert-defined question types with type-specific evaluation metrics.

\item We propose VaseVL, a model developed through a two-stage training process with taxonomy-aware reward shaping, designed to better leverage the expert annotations provided by VaseVQA beyond surface-level visual features.

\item Experimental results show that while SFT plateaus in reasoning on this dataset, our targeted RL approach consistently improves performance on challenging question types, highlighting the value of high-quality domain-specific data for expert-level cultural heritage analysis.

\end{itemize}

\section{Related Work}

\paragraph{Multimodal Large Language Models (MLLMs).}
Pretrained vision–language models learn joint representations from large-scale multimodal corpora and have advanced a wide range of tasks, including image–text retrieval, VQA, grounding, and dialogue \cite{li2019visualbert, chen2020uniter, X-VLM, li2021align,song2025audio,song2025geolocation,song2025maniplvm,huang20253d,liu2025nav}. Building on this, Visual Instruction Tuning (VIT) further enhances MLLMs’ ability to understand and execute complex multimodal instructions, as exemplified by LLaVA \cite{llava}, MiniGPT-4 \cite{minigpt4}, Gemini~1.5 \cite{Gemini1.5}, and Qwen2.5-VL-Instruct \cite{qwen2.5-VL}.
As shown in Table~\ref{tab.eval_datasets} in the appendix, current MLLMs are typically evaluated across a broad suite of vision tasks, validating their general perception and cross-modal alignment capabilities. However, such generic evaluations remain insufficient for highly specialized scenarios such as cultural-heritage analysis, which require integrating fine-grained visual cues with historically and archaeologically grounded knowledge to support expert-level reasoning and judgment.

\paragraph{VQA Benchmarks.}
Foundational VQA datasets, including VQA \cite{VQA}, COCO-QA \cite{COCO-QA}, and Visual7W \cite{Visual7W}, enabled rapid progress but primarily cover generic objects and scenes. In cultural-heritage domains, publicly available resources remain scarce. RePAIR \cite{RePAIR} targets oracle bones, and HUST-OBS \cite{HUST-OBS} focuses on fragment reconstruction, leaving a gap for classical artifacts such as ancient Greek vases. Table~\ref{tab:major_VQA_datasets} summarizes the key characteristics of major VQA and visual reasoning datasets. Our VaseVQA fills this gap with a VQA-centric benchmark designed to probe factual recall (e.g., \emph{Fabric}, \emph{Technique}) and expert-level reasoning (\emph{Attribution}, \emph{Decoration}, \emph{Date}, \emph{Provenance}, \emph{Shape}) under a type-aware evaluation protocol.

\begin{table*}[ht]
  \begin{minipage}{0.99\textwidth}
    \centering  
    \vspace{-4mm}
    \small   %
\begin{tabular}{p{2cm} p{12.5cm}}
\toprule
 \multicolumn{2}{l}{\bf VaseVQA example:}  \\
\midrule
&  \includegraphics[height=2.5cm]{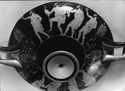} \\
\midrule
Question  & What is the fabric of the vase? \\
Answer & The fabric of the vase is ATHENIAN. \\
\midrule
Question & What is the technique of the vase? \\
Answer & The technique of the vase is RED-FIGURE. \\
\midrule
Question & What is the shape name of the vase? \\
Answer & The shape name of the vase is CUP B. \\
\midrule
Question & What is the provenance of the vase? \\
Answer & The provenance of the vase is not available. \\
\midrule
Question & What is the date of the vase? \\
Answer & The date of the vase is -450 to -400. \\
\midrule
Question & What is the attribution of the vase? \\
Answer & The vase is attributed to CODRUS P by BURN | CODRUS P by UNKNOWN. \\
\midrule
Question & What is the decoration of the vase? \\
Answer & The decoration of the vase is A,B: THEATRICAL, DRAPED SATYRS, WITH STORK, BOX AND SANDAL, ARYBALLOI, OINOCHOE, LYRE AND STAFFS, ONE CONFRONTING DRAPED YOUTH | I: AMAZON ON HORSEBACK. \\
\bottomrule
\end{tabular}
\vspace{-1mm}
\caption{\textbf{Example of a VaseVQA Dataset}: Eight attribute-specific questions (fabric, technique, shape, provenance, date, attribution, decoration) paired with visual input, presented in a conversational Q\&A format to analyze an Athenian red-figure cup (450–400 BCE) attributed to the Codrus Painter.}
\label{tab:visual_example}  
  \end{minipage}
  \vspace{-3mm}
\end{table*}

\section{Data Collection}

VaseVQA was constructed with a focus on representing Ancient Greek culture, ensuring both the visual and textual components accurately reflect the region's rich cultural heritage. The Table~\ref{tab:major_VQA_datasets} presents major VQA and Visual Reasoning datasets, compared with the VaseVQA. Table~\ref{tab:splits} shows the division of the VaseVQA dataset into training and test sets, with each image having 7 questions. In addition, Table~\ref{tab:visual_example} illustrates examples within the dataset, demonstrating the questions crafted for every sample.

\subsection{Image Collection}

The images in this dataset were collected through collaborations with Ancient Greek archaeological institutions, museums, and cultural heritage centers~\cite{carc_beazley_pottery}. 
We focused on gathering images of classical funerary vases that are commonly found in Ancient Greek archaeological sites, ensuring a diverse representation of artifacts across different cultural groups. The images include both complete objects and fragments, as well as images of the vases in their original burial contexts. This collection was designed to capture intricate details of materials, craftsmanship, and regional variations on the Table \ref{tab:visual_example}.

\subsection{Text Collection}
The textual data for the dataset was derived from several key sources, including academic papers, archaeological reports, and expert annotations provided by Ancient Greek historians and cultural heritage experts. The texts are descriptions of the artifacts, detailing their material composition (such as red pottery or glazed ceramics), motifs (such as human, animal, or abstract designs), and the archaeological context (such as burial sites or ceremonial uses). These descriptions were translated and structured to align with the images and make the data accessible for vision-language tasks.

\subsection{Annotation}
The dataset was labeled by a team of archaeologists and cultural heritage experts, who annotated the images with several key attributes. These include the material (e.g., red pottery, glazed ceramics), pattern type (e.g., human figures, animal motifs, abstract symbols), excavation layer, radiocarbon dating estimates, manufacturing techniques (e.g., hand-built, wheel-thrown, firing temperature), and the contextual use of the object (e.g., funerary or ceremonial). The experts also identified restoration marks, if applicable. The labeling process ensures a high level of detail and accuracy in reflecting the cultural and historical context of each artifact.

\begin{table*}[!t]
\centering
\setlength{\tabcolsep}{10pt}
\resizebox{\textwidth}{!}{
\begin{tabular}{l c c c c c c c}
\toprule
\textbf{Split} & \textbf{Fabric} & \textbf{Technique} & \textbf{Shape} & \textbf{Provenance} & \textbf{Date} & \textbf{Attribution} & \textbf{Decoration} \\
\midrule
\multicolumn{8}{l}{\textit{Question Type Distribution (\%)}} \\
Train & 17.3 & 17.3 & 17.3 & 6.7 & 17.3 & 7.0 & 17.1 \\
Test  & 17.2 & 17.2 & 17.2 & 6.9 & 17.2 & 7.1 & 17.1 \\
\midrule
\multicolumn{8}{l}{\textit{Average Answer Length (words)}} \\
Avg. & 10.0 & 11.0 & 13.0 & 16.6 & 12.0 & 20.9 & 28.3 \\
\midrule
\textbf{Total} & \multicolumn{7}{c}{ \quad 67,614 Question--Answer Pairs} \\
\bottomrule
\end{tabular}
}
\vspace{-1mm}
\caption{
Dataset composition of VaseVQA.
The dataset is split into training and test sets with a 4:1 ratio.
Question-type distributions are nearly identical across splits.
Attribution and Decoration exhibit longer average answers, reflecting their more open-ended response requirements.
}
\label{tab:splits}
\end{table*}

\section{Methods}

We start from a supervised fine-tuned (SFT) model trained on the ancient Greek vase dataset. While the model demonstrates basic visual question-answering capability, it exhibits systematic weaknesses in reasoning and compositional understanding. We apply reinforcement learning to address these limitations, enhancing reasoning ability while preserving foundational knowledge.

\begin{figure*}[t]
    \centering
\includegraphics[width=\linewidth]{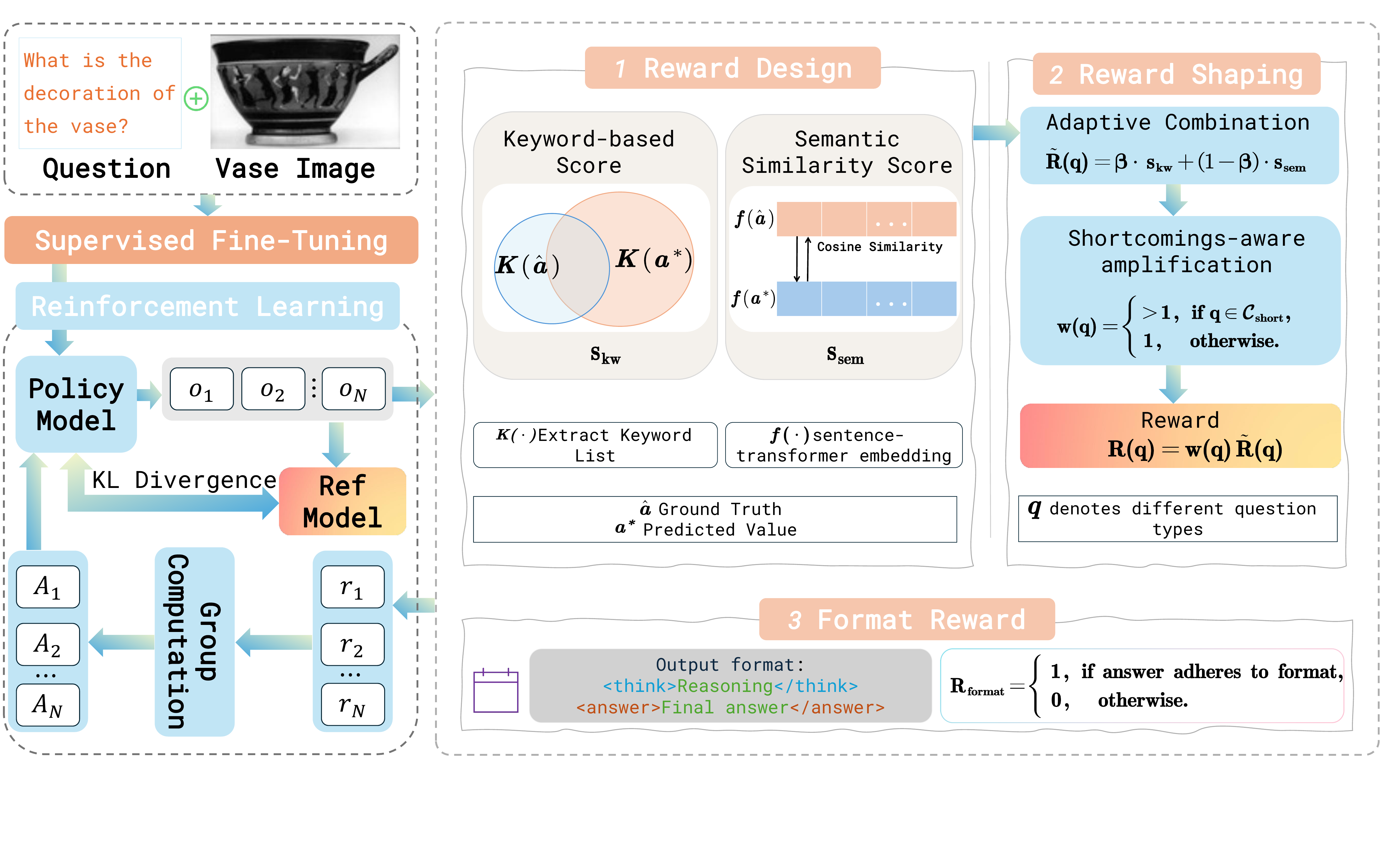}
    \caption{Overall framework of VaseVL. The proposed pipeline integrates SFT with RL under the GRPO paradigm. Given a vase image $x$, a question $q$, and the reference answer $a^*$, the model refines its reasoning ability by balancing lexical and semantic rewards while constraining policy drift from $\pi_{\text{ref}}$.}
    \label{fig:vasevl_framework}
    \vspace{-3mm}
\end{figure*}

We denote the post-SFT model as the reference policy and initialize a trainable policy from it. The objective is to achieve expert-level accuracy and compositional robustness. To ensure stable training across diverse question types, we employ Group Relative Policy Optimization (GRPO). For each image-question pair, we sample multiple candidate answers and compute type-conditioned rewards. GRPO normalizes these rewards per prompt by subtracting the average, yielding relative advantages that are robust to cross-type scale variation. We then optimize a clipped PPO-style objective with KL regularization to prevent drift from the reference policy (Figure~\ref{fig:vasevl_framework}).

To make GRPO effective, we design a reward engineering framework tailored to the shortcomings of the SFT model. The reward consists of two complementary metrics: (1) a keyword-based score measuring lexical overlap, prioritizing factual correctness, and (2) a semantic similarity score based on normalized cosine similarity of sentence embeddings. Recognizing that different tasks weigh precision versus semantics differently, we combine them with adaptive, type-conditioned weights. Furthermore, for error-prone categories, we amplify the reward signal with an additional weighting factor, focusing learning on areas where the SFT model is weakest.

\subsection{Reward Design}  

To effectively guide model optimization, we propose a comprehensive reward function that directly addresses the limitations of the SFT model and flexibly adapts to different categories of questions. The reward integrates both lexical and semantic perspectives, ensuring that generated answers $\hat{a}$ are evaluated not only for factual precision but also for meaning preservation with respect to the ground-truth reference $a^*$.  

\paragraph{Keyword-based Score ($s_{\text{kw}}$):}  
Lexical accuracy is particularly important for factual and knowledge-intensive queries. To capture this, we define a keyword-based score that measures the overlap of essential terms between the model output and the reference:

\vspace{-5mm}
\begin{equation}
    s_{\text{kw}} = \frac{|K(\hat{a}) \cap K(a^*)|}{|K(\hat{a}) \cup K(a^*)|},
\end{equation}

where $K(\cdot)$ extracts the keyword set from a given text. This formulation is analogous to a Jaccard similarity over keywords, prioritizing factual correctness and penalizing omissions or hallucinated entities.  

\paragraph{Semantic Similarity Score ($s_{\text{sem}}$):}  
While keyword overlap captures precision, it may fail to reflect semantic equivalence when paraphrases are used. To complement this, we compute semantic similarity using the \texttt{all-MiniLM-L6-v2} model from the Sentence-Transformers library. Each text string is embedded into a high-dimensional vector space, and the similarity is defined as the normalized cosine similarity:

\vspace{-3mm}
\begin{equation}
    s_{\text{sem}} = \frac{\cos(f(\hat{a}), f(a^*)) + 1}{2}, 
\end{equation}

where $f(\cdot)$ denotes the embedding function from~\citep{wang2020minilm}. This design ensures that semantically faithful answers, even if phrased differently, receive appropriate credit.

\subsection{Reward Shaping}  

A critical insight is that the importance of lexical accuracy versus semantic equivalence is not uniform across tasks. For example, in factual questions such as \textit{``What is the capital of Canada?''}, exact keyword matching (``Ottawa'') is indispensable. In contrast, for descriptive or open-ended queries like \textit{``Explain the impact of climate change''}, semantic similarity plays a more dominant role, as diverse phrasings can still convey correct meaning.  

To incorporate this adaptivity, we define the reward as a weighted combination:  

\vspace{-5mm}
\begin{equation}
    \tilde{R}(q) = \beta(q) s_{\text{kw}} + (1 - \beta(q)) s_{\text{sem}},
\end{equation}

where the weight $\beta(q)$ is conditioned on the question type $q$.
This weight allows the reward to adapt its emphasis to the requirements of different tasks.

Furthermore, to emphasize learning in areas where the SFT model is weakest, we apply an additional scaling factor:  

\vspace{-5mm}
\begin{equation}
    R(q) = w(q) \cdot \tilde{R}(q),
\end{equation}

with  

\vspace{-5mm}
\begin{equation}
    w(q) = 
    \begin{cases} 
    > 1, & \text{if } q \in \mathcal{C}_{\text{short}}; \\
    1, & \text{otherwise.}
    \end{cases}
\end{equation}

Here, $\mathcal{C}_{\text{short}}$ denotes a predefined set of question types where the SFT model is known to underperform (e.g., numerical reasoning, multi-step factual queries). By amplifying the reward signal in these regions, the training process allocates greater attention to challenging categories, accelerating targeted improvements without neglecting overall performance.

\section{Experiments}
In this section, we conduct a series of experiments to validate the effectiveness of our proposed model, VaseVL. We evaluate our model on the newly introduced VaseVQA benchmark and compare its performance against various strong baselines, including general-purpose Multimodal Large Language Models (MLLMs) and a fine-tuned version of the model without reinforcement learning. Our evaluation is structured around the taxonomy of seven distinct question types to provide a granular analysis of the model's capabilities in expert-level reasoning.

\begin{table*}[t]
    \hspace{-3mm}
	\resizebox{\textwidth}{!}{
	\begin{tabular}{l | c | cccccc | c | c}
        \toprule[1pt]

		\multirow{2}{*}{Model} & \multirow{2}{*}{Param.} & Fabric & Technique & Shape & Provenance & Date & Attribution & \multicolumn{1}{c|}{Decoration} \\
        \cmidrule(r){3-8} \cmidrule(r){9-10}

         & & \multicolumn{6}{c}{Accuracy} & \multicolumn{1}{|c|}{BLEU@1} & Overall \\

        \midrule
LLaVA~\cite{llava} & 7B & 11.56 & 0 & 44.60 & 0 & 0 & 28.41 & 6.28 & 14.10 \\
Vicuna~\cite{zheng2023judging} & 7B & 0 & 0 & 0.24 & 0 & 0 & 0 & 1.14 & 0.04 \\
MiniCPM~\cite{yao2024minicpm} & 8B & 0.29 & 0.03 & 0 & 0.97 & 0 & 0.14 & 1.15 & 0.24 \\
InternVL2~\cite{chen2024far} & 1B & 10.50 & 0.08 & 3.97 & 11.52 & 0 & 14.10 & 3.41 & 6.70 \\
InternVL2~\cite{chen2024far} & 2B & 0 & 0.60 & 6.03 & 8.62 & 0.65 & 3.99 & 1.94 & 3.31 \\
InternVL2~\cite{chen2024far} & 8B & 1.69 & 4.00 & 7.66 & 21.24 & 0 & 2.85 & 2.00 & 6.24 \\
Qwen-2.5-VL~\cite{qwen2.5-VL} & 3B & 0.29 & 0.02 & 0.14 & 0.00 & 14.86 & 0.00 & 2.29 & 2.55 \\
Qwen-2.5-VL~\cite{qwen2.5-VL} & 7B & 0.00 & 0.00 & 0.05 & 0.00 & 17.95 & 0.00 & 1.91 & 3.00 \\
       \midrule
Gemini-2.0-Flash & -- & 9.37 & 90.11 & 26.22 & 12.42 & 17.41 & 2.92 & 6.79 & 26.41 \\

GPT-4o-mini & -- & 80.21 & 86.08 & 76.95 & 68.84 & \textbf{47.22} & 39.51 & 6.53 & 66.47 \\
       \midrule
\textbf{VaseVL (Ours)} & 3B & \textbf{99.95} & \textbf{95.93} & \textbf{83.99} & \textbf{73.67} & 39.87 & \textbf{60.83} & \textbf{9.82} & \textbf{75.71} \\

    \bottomrule[1pt]
	\end{tabular}
	}
	\caption{Performance comparison on the VaseVQA benchmark.}
	\label{tab:main-result0}
\end{table*}

\subsection{Implementation details}
\label{sec:sft}

We propose a two-stage training paradigm. In the first stage, we perform full-parameter supervised fine-tuning (SFT) on the MLLMs, and in the second stage, we conduct task-specific reinforcement learning (RL) training. Concretely, we initialize from a general-purpose MLLM and instruction-tune on $\mathcal{D}=\{(x_i,q_i,a_i^{*})\}_{i=1}^N$.

The SFT model $\pi_{\text{ref}}$ is used both as a stable reference for RL and as a probe to evaluate per-type performance over $\mathcal{T}$ (e.g., \emph{Fabric}, \emph{Technique}), from which we select a shortcoming subset $\mathcal{C}_{\text{short}}\subseteq\mathcal{T}$ for targeted improvement. For SFT, we adopt a per-device batch size of $1$ with $8{\times}$ gradient accumulation, a cosine learning rate schedule with initial value $1{\times}10^{-4}$, one epoch of training, a warmup ratio of $0.1$, and enable \texttt{bf16}. After SFT, we perform RL focused on $\mathcal{C}_{\text{short}}$ with a GRPO-style setup: $8$ rollouts per prompt, temperature $0.9$, one iteration per batch with KL penalty coefficient $0.04$, training for $2$ epochs at learning rate $1{\times}10^{-6}$.

\subsection{Evaluation Metrics}
We use a task-specific evaluation protocol, applying the most appropriate metric to each question type. We report accuracy for factual question types, with task-specific computation protocols.

\paragraph{ANLS-based Accuracy}
For short-answer factual questions—\textit{Fabric}, \textit{Technique}, \textit{Shape}, \textit{Provenance}, and \textit{Attribution}—we compute accuracy using Average Normalized Levenshtein Similarity (ANLS), a ST-VQA–inspired soft metric robust to minor character-level and OCR-like errors~\cite{biten2019stvqa}.

\paragraph{Date Accuracy}
For \textit{Date} questions, we parse numerical years and define accuracy as the maximum over three components: partial credit for correct date-range formatting, a proximity-based score within a tolerance margin, and a primary score based on the Intersection over Union (IoU) of year ranges.

\paragraph{BLEU@1 Score}
For the descriptive \textit{Decoration} questions, the target output consists of a list of specific visual attributes and keywords, with an emphasis on lexical precision. Accordingly, we adopt BLEU@1 as the evaluation metric, measuring unigram precision between the generated outputs and the ground truth. This metric effectively captures key descriptive terms without overly penalizing stylistic variations, and serves as a proxy for the model’s compositional understanding.

\subsection{Main Results}

\paragraph{Effect of Model Scale}
The results in Table~\ref{tab:main-result0} indicate that simply increasing model size does not lead to performance gains on VaseVQA. Larger general-purpose MLLMs, such as LLaVA 7B~\cite{llava}, Vicuna 7B~\cite{zheng2023judging}, and MiniCPM 8B~\cite{yao2024minicpm}, fail to outperform smaller models and in some expert-level tasks even achieve scores close to zero. This outcome is primarily due to the lack of cultural-heritage knowledge in their pretraining data: ancient Greek pottery styles, shapes, and historical contexts are almost absent from web-scale corpora. In contrast, VaseVL, with only 3B parameters, integrates supervised fine-tuning to supplement domain knowledge and diagnosis-guided reinforcement learning to activate reasoning. This combination allows our model to surpass all larger baselines. The result underscores that domain-specific data and training strategies are more critical than raw model scale, enabling smaller, efficient models to achieve expert-level performance.

\begin{table*}[t]
\small
\resizebox{\textwidth}{!}{
\begin{tabular}{l c | cccccc | cc | c}
\toprule[1pt]

\multirow{2}{*}{SFT} & \multirow{2}{*}{RL} 
& \multicolumn{6}{c}{Fabric \quad Technique \quad Shape \quad Provenance \quad Date \quad Attribution}
& \multicolumn{2}{|c|}{Decoration} 
& \multicolumn{1}{c}{} \\
\cmidrule(r){3-8} \cmidrule(r){9-10} \cmidrule(r){11-11}

& & \multicolumn{6}{c}{Accuracy} 
&  \multicolumn{1}{|c}{BLEU@1} & Prometheus Score & Overall\\
\midrule
- & - & 0.29 & 0.02 & 0.14 & 0.00 & 14.86 & 0.00 & 2.29 & 0.99 & 2.55 \\
- & \checkmark & 13.33 & 19.95 & 14.82 & 5.27 & 3.58 & 11.50 & 4.82 & 2.91 & 11.41 \\
\checkmark & - & \textbf{99.96} & 94.99 & 83.98 & 71.67 & 37.96 & 56.96 & 2.57 & 10.31 & 74.25 \\
\checkmark & \checkmark & 99.95 & \textbf{95.93} & \textbf{83.99} & \textbf{73.67} & \textbf{39.87} & \textbf{60.83} & \textbf{9.82} & \textbf{14.71} & \textbf{75.71} \\
\bottomrule[1pt]
\end{tabular}
}
\vspace{-1mm}
\caption{Ablation study results. The final row reports the performance of our proposed VaseVL. Prometheus Score is the LLM-as-a-judge score produced by \textit{Prometheus-v2.0-7B}.}
\vspace{2mm}
\label{tab:main-result}
\end{table*}

\begin{table*}[t]
    \hspace{-3mm}
	\resizebox{\textwidth}{!}{
	\begin{tabular}{l | cccccc | c | c}
        \toprule[1pt]

		\multirow{2}{*}{VaseVQA-mini} 
        & Fabric & Technique & Shape & Provenance & Date & Attribution 
        & \multicolumn{1}{c|}{Decoration} \\
        \cmidrule(r){2-7} \cmidrule(r){8-9}
         & \multicolumn{6}{c}{Accuracy} & \multicolumn{1}{|c|}{BLEU@1} & Overall \\
        \midrule

        w/o Keyword ($s_{\text{kw}}$)      
        & 99.95 & 94.62 & 83.43 & 72.27 & 38.13 & 57.14 & 5.93 & 74.25 \\

        w/o Semantic ($s_{\text{sem}}$)    
        & 99.95 & 94.83 & 83.74 & 72.86 & 37.53 & 56.94 & 3.24 & 74.30 \\

        w/o Amplification ($w(q)$)         
        & 99.96 & \textbf{94.97} & \textbf{84.02} & 73.05 & 38.04 & 57.23 & 4.13 & 74.54 \\

        \textbf{VaseVL}                        
        & \textbf{99.96} & 94.93 & 83.98 & \textbf{73.24} & \textbf{38.55} & \textbf{57.79} & \textbf{6.21} & \textbf{74.74} \\

        \bottomrule[1pt]
	\end{tabular}
	}
    \vspace{-1mm}
	\caption{Performance comparison on the VaseVQA benchmark.}
    \vspace{-2mm}
	\label{tab:main-result1}
\end{table*}

\paragraph{Task-Type Variability}
Low-level tasks such as \emph{Fabric} and \emph{Technique} fail in zero-shot settings due to missing domain-specific textures, but reach near-perfect accuracy after supervised fine-tuning, indicating a reliance on visual feature coverage rather than explicit reasoning.
Mid-level tasks including \emph{Shape} and \emph{Provenance} exhibit limited zero-shot capability, likely inherited from analogous patterns in pretraining data, yet achieving expert-level performance requires domain alignment through supervised fine-tuning and reinforcement learning.
High-level tasks—\emph{Date}, \emph{Attribution}, and \emph{Decoration}—expose fundamental limitations of current VLMs: zero-shot models largely collapse, and although VaseVL attains substantial improvements, with 39.87\% on \emph{Date}, 60.83\% on \emph{Attribution}, and 9.82 BLEU@1 on \emph{Decoration}, a significant gap to easier tasks remains.
These tasks demand historical reasoning, multi-evidence integration, and compositional language generation beyond pure visual recognition. Overall, low-level perception adapts rapidly with data exposure, mid-level reasoning benefits from domain alignment, while high-level historical reasoning remains a core bottleneck in current vision–language modeling.

\subsection{Ablation Study}

\paragraph{Effect of Training Stages} 
Table~\ref{tab:main-result} compares the contributions of Supervised Fine-Tuning (SFT) and Reinforcement Learning (RL). 
The zero-shot baseline's low overall score (2.55) highlights the domain gap in general-purpose models. 
RL without prior SFT yields only marginal gains in reasoning (e.g., 11.50\% on \textit{Attribute}), lacking sufficient grounding. 
Conversely, SFT-only excels in factual recall (\textit{Fabric}: 99.96\%, \textit{Technique}: 94.99\%) but struggles with complex generation, evidenced by low scores on \textit{Date} (37.96\%) and \textit{Decoration} (2.57 BLEU@1). 
VaseVL integrates both to achieve superior performance (75.71\% overall), retaining factual accuracy while significantly boosting reasoning (e.g., \textit{Decoration} improves to 9.82 BLEU@1). 
Additional evaluation using Prometheus-7B-v2.0~\cite{kim-etal-2024-prometheus} as an LLM-judge further corroborates the efficacy of this two-stage strategy in enhancing descriptive quality.

\paragraph{Effect of Reward Components} 
We analyze reward efficacy on \textit{VaseVQA-mini} (a stratified 20\% subset) in Table~\ref{tab:main-result1}.
Ablating the Keyword Score ($s_{\text{kw}}$) impairs strict classification (\textit{Technique} drops to 94.62\%, \textit{Shape} to 83.43\%), confirming the need for lexical precision. 
Removing the Semantic Score ($s_{\text{sem}}$) degrades open-ended generation, with \textit{Decoration} plummeting from 6.21 to 3.24 BLEU-1, indicating the importance of embedding-based guidance. 
Finally, excluding the Shortcomings-aware Amplification ($w(q)$) hinders reasoning on difficult tasks like \textit{Date} (38.04\%) and \textit{Attribution} (57.23\%). 
The full VaseVL framework achieves the highest overall score (74.74\%), validating the synergistic value of these reward signals.

\section{Conclusion}
We introduced VaseVL, a model trained with a two-stage framework and reward shaping for cultural heritage analysis, together with VaseVQA, a comprehensive benchmark spanning seven question types. By aligning optimization with task-specific evaluation metrics and amplifying rewards for complex reasoning, VaseVL moves beyond surface-level pattern recognition. Specifically, GRPO with KL regularization preserves factual accuracy while enabling expert-level reasoning. The release of VaseVQA establishes a reproducible standard for the field. Our findings on ancient Greek pottery indicate that reinforcement learning with targeted reward shaping is a promising direction for adapting MLLMs to other domains that require expert-level reasoning. Future work will validate this approach on a broader range of cultural heritage datasets.

\section{Social Impact}

VaseVL, a foundational MLLM for Ancient Greek pottery, preserves and interprets this vital cultural heritage by integrating visual and textual analysis of styles, shapes, decorations, and inscriptions. It aids archaeologists in accurate classification and digital documentation, enhances access to global collections, and supports forgery detection and cultural preservation. Museums and educators can also use VaseVL to create engaging learning experiences. Ultimately, VaseVL bridges AI and archaeology, safeguarding the historical legacy of Ancient Greek pottery for future generations.

\section{Limitations}

Although VaseVQA provides a large-scale benchmark with carefully designed training and test splits, data bias and coverage remain important limitations. The dataset focuses on ancient Greek pottery, and while it captures a wide range of fabrics, techniques, shapes, and decorative styles, it cannot fully represent the diversity of cultural heritage artifacts across different regions, periods, and materials. In addition, question–answer distributions may reflect annotation preferences and existing scholarly conventions, which could inadvertently bias model training. As a result, models optimized on VaseVQA may overfit to dataset-specific patterns rather than generalize to real-world archaeological or art-historical contexts. Future work should address these issues by incorporating more heterogeneous artifacts, broader cultural traditions, and multiple sources of expert annotation to improve robustness and reduce bias.

\bibliography{custom}
\newpage

\section*{Appendix}

Current MLLMs are typically evaluated under a broad suite of visual tasks, including image classification, image–text retrieval, object detection, semantic segmentation, and general-purpose VQA, as summarized in Table~\ref{tab.eval_datasets}.

\begin{table*}
\centering
    \small
    \renewcommand\arraystretch{1}
    \begin{tabular}{p{1.5cm}|p{5.2cm}|c|p{0.8cm}|c|c|p{2.4cm}}
        \toprule[1pt]
        \textbf{Task}&\textbf{Dataset} & \textbf{Year} &\textbf{Classes} & \textbf{Training} & \textbf{Testing} & \textbf{Evaluation Metric} \\
        \midrule
        Image  &MNIST~\cite{lecun1998gradient}~\href{http://yann.lecun.com/exdb/mnist/}{[link]} &1998& 10 & 60,000 & 10,000 & Accuracy \\
        Classification &Caltech-101~\cite{fei2004learning}~\href{https://data.caltech.edu/records/mzrjq-6wc02}{[link]} & 2004& 102 & 3,060 & 6,085 & Mean Per Class\\
        &PASCAL VOC 2007 Classification~\cite{everingham2010pascal}~\href{http://host.robots.ox.ac.uk/pascal/VOC/voc2007/}{[link]} & 2007 & 20 & 5,011 & 4,952 & 11-point mAP \\
        &Oxford 102 Folwers~\cite{nilsback2008automated}~\href{https://www.robots.ox.ac.uk/~vgg/data/flowers/102/}{[link]} &2008& 102 & 2,040 & 6,149 & Mean Per Class\\

        &ImageNet-1k~\cite{deng2009imagenet}~\href{https://www.image-net.org/}{[link]} &2009& 1000 & 1,281,167 & 50,000 & Accuracy\\
        &SUN397~\cite{xiao2010sun}~\href{https://vision.princeton.edu/projects/2010/SUN/}{[link]} & 2010& 397 & 19,850 & 19,850 & Accuracy\\
        &SVHN~\cite{netzer2011reading}~\href{http://ufldl.stanford.edu/housenumbers/}{[link]} &2011&10 &73,257&26,032&Accuracy\\
        &STL-10~\cite{coates2011analysis}~\href{https://cs.stanford.edu/~acoates/stl10/}{[link]} &2011& 10 & 1,000 & 8,000 & Accuracy\\
        &GTSRB~\cite{stallkamp2011german}~\href{https://www.kaggle.com/datasets/meowmeowmeowmeowmeow/gtsrb-german-traffic-sign}{[link]} &2011& 43 & 26,640 & 12,630 & Accuracy\\
        &KITTI Distance~\cite{geiger2012we} ~\href{https://github.com/harshilpatel312/KITTI-distance-estimation}{[link]}& 2012& 4 & 6,770 & 711 & Accuracy\\
        &IIIT5k~\cite{mishra2012scene}~\href{https://cvit.iiit.ac.in/research/projects/cvit-projects/the-iiit-5k-word-dataset}{[link]} &2012&36 &2,000&3,000&Accuracy\\
        &Oxford-IIIT PETS~\cite{parkhi2012cats}~\href{https://www.robots.ox.ac.uk/~vgg/data/pets/}{[link]} &2012 & 37 & 3,680 & 3,669 & Mean Per Class\\

        &FGVC Aircraft~\cite{maji2013fine}~\href{https://www.robots.ox.ac.uk/~vgg/data/fgvc-aircraft/}{[link]} &2013 & 100 & 6,667 & 3,333 & Mean Per Class\\
        &Facial Emotion Recognition 2013~\cite{goodfellow2013challenges}~\href{https://www.kaggle.com/competitions/challenges-in-representation-learning-facial-expression-recognition-challenge/data}{[link]} &2013& 8 & 32,140 & 3,574 & Accuracy\\
        &Rendered SST2~\cite{socher2013recursive}~\href{https://github.com/openai/CLIP/blob/main/data/rendered-sst2.md}{[link]} &2013& 2 & 7,792 & 1,821 & Accuracy\\

        &Describable Textures (DTD)~\cite{cimpoi2014describing}~\href{https://www.robots.ox.ac.uk/~vgg/data/dtd/}{[link]} & 2014 & 47 & 3,760 & 1,880 & Accuracy\\
        &Food-101~\cite{bossard2014food}~\href{https://www.kaggle.com/datasets/dansbecker/food-101}{[link]} & 2014 & 102 & 75,750 & 25,250 & Accuracy  \\
        &Birdsnap~\cite{berg2014birdsnap}~\href{https://thomasberg.org/}{[link]} & 2014 & 500 & 42,283 & 2,149 & Accuracy\\
        &RESISC45~\cite{cheng2017remote}~\href{https://pan.baidu.com/s/1mifR6tU?_at_=1679281159364#list/path=\%2F}{[link]} &2017& 45 & 3,150 & 25,200 & Accuracy\\
        &CLEVR Counts~\cite{johnson2017clevr}~\href{https://cs.stanford.edu/people/jcjohns/clevr/}{[link]} &2017& 8 & 2,000 & 500 & Accuracy\\
        &PatchCamelyon~\cite{veeling2018rotation}~\href{https://github.com/basveeling/pcam}{[link]} &2018& 2 & 294,912 & 32,768 & Accuracy\\

        &EuroSAT~\cite{helber2019eurosat}~\href{https://github.com/phelber/eurosat}{[link]} &2019& 10 & 10,000 & 5,000 & Accuracy\\
        
        &Hateful Memes~\cite{kiela2020hateful}~\href{https://ai.facebook.com/blog/hateful-memes-challenge-and-data-set/}{[link]} & 2020&2 & 8,500 & 500 & ROC AUC\\
        &Country211~\cite{radford2021learning}~\href{https://github.com/openai/CLIP/blob/main/data/country211.md}{[link]} &2021& 211 & 43,200 & 21,100 & Accuracy\\

        \midrule
        Image-Text &Flickr30k~\cite{young2014image}~\href{https://shannon.cs.illinois.edu/DenotationGraph/}{[link]} &2014& - &31,783&-&Recall\\
        Retrieval &COCO Caption~\cite{chen2015microsoft}~\href{https://github.com/tylin/coco-caption}{[link]}&2015 & - &82,783&5,000 &Recall\\
        \midrule
        
        Action &UCF101~\cite{soomro2012ucf101}~\href{https://www.crcv.ucf.edu/data/UCF101.php}{[link]} &2012& 101 & 9,537 & 1,794 & Accuracy\\
        Recognition &Kinetics700~\cite{carreira2019short}~\href{https://www.deepmind.com/open-source/kinetics}{[link]} &2019& 700 & 494,801 & 31,669 & Mean(top1, top5)\\
        &RareAct~\cite{miech2020rareact}~\href{https://github.com/antoine77340/RareAct}{[link]} &2020&122 &7,607&-&mWAP, mSAP\\

        \midrule
        
        Object Detection&COCO 2014 Detection~\cite{lin2014microsoft}~\href{https://www.kaggle.com/datasets/jeffaudi/coco-2014-dataset-for-yolov3}{[link]} & 2014& 80 & 83,000 & 41,000 & box mAP \\
        &COCO 2017 Detection~\cite{lin2014microsoft}~\href{https://www.kaggle.com/datasets/awsaf49/coco-2017-dataset}{[link]} & 2017 & 80 & 118,000 & 5,000 & box mAP \\
        &LVIS~\cite{gupta2019lvis}~\href{https://www.lvisdataset.org/}{[link]}& 2019& 1203 & 118,000 &5,000& box mAP \\
        &ODinW~\cite{li2022elevater}~\href{https://eval.ai/web/challenges/challenge-page/1839/overview}{[link]} &2022& 314 & 132413 & 20070 & box mAP \\ 
        \midrule
        Semantic Segmentation&PASCAL VOC 2012 Segmentation~\cite{everingham2010pascal}~\href{http://host.robots.ox.ac.uk/pascal/VOC/voc2012/}{[link]} & 2012&20&1464&1449&mIoU\\
        &PASCAL Content~\cite{mottaghi2014role}~\href{https://www.cs.stanford.edu/~roozbeh/pascal-context/}{[link]} & 2014 &459&4998&5105 &mIoU\\
        &Cityscapes~\cite{cordts2016cityscapes}~\href{https://www.cityscapes-dataset.com/}{[link]}& 2016 &19&2975&500&mIoU \\
        &ADE20k~\cite{zhou2017scene}~\href{https://groups.csail.mit.edu/vision/datasets/ADE20K/}{[link]}&2017&150& 25574 & 2000 & mIoU \\
        \midrule
        Vase & VaseVQA & 2025 & 7 & 9534 & 2339 & Accuracy \& BLEU \\
        \bottomrule[1pt]
    \end{tabular}
    \caption{Summary of the widely-used visual recognition datasets for MLLM evaluation.}
        \label{tab.eval_datasets}
\end{table*}

\end{document}